\documentclass{article}

\usepackage{times}
\usepackage{graphicx}
\usepackage{subfigure}

\usepackage{amssymb}

\usepackage{natbib}

\usepackage{algorithm}
\usepackage{algorithmic}

\usepackage{hyperref}

\usepackage[arxiv]{icml2017}

\usepackage{amsmath}

\usepackage{capt-of}

\begin{document} 

\footskip 31pt

\twocolumn[
    \icmltitle{Improving the Neural GPU Architecture for Algorithm Learning}    

    
    
    
    \begin{icmlauthorlist}
    \icmlauthor{Karlis Freivalds}{mii}
    \icmlauthor{Renars Liepins}{mii}
    \end{icmlauthorlist}
    
    \icmlaffiliation{mii}{Institute of Mathematics and Computer Science University of Latvia, Raina bulvaris 29, Riga, LV-1459, LATVIA}
    
    \icmlcorrespondingauthor{Karlis Freivalds}{karlis.freivalds@lumii.lv}    

    \icmlkeywords{algorithm learning, data efficiency, neural gpu}    
    \vskip 0.3in
]


\printAffiliationsAndNotice{}  


\begin{abstract} 
Algorithm learning is a core problem in artificial intelligence with significant implications on automation level that can be achieved by machines.
Recently deep learning methods are emerging for synthesizing an algorithm from its input-output examples, the most successful being the Neural GPU, capable of learning multiplication. We present several improvements to the Neural GPU that substantially reduces training time and improves generalization. 
We introduce a new technique - hard nonlinearities with saturation costs - that has general applicability.
We also introduce a technique of diagonal gates that can be applied to active-memory models. 
The proposed architecture is the first capable of learning decimal multiplication end-to-end.

\end{abstract}


\section{Introduction}
\label{introduction}

Deep Neural Networks have achieved state-of-the-art results in a wide range of tasks, notably in computer vision \cite{krizhevsky2012imagenet}, speech recognition \cite{amodei2015deep}, and natural language processing \cite{bahdanau2014neural} and extensive work is performed to broaden its scope of application. 
A particularly interesting area is algorithm learning i.e. synthesizing an algorithm from its input-output examples. It is a long-standing open problem with theoretical results dating back to \cite{gold1967language, angluin1983inductive} but still, the results are far from reaching an industry scale.

With the emergence of powerful deep learning methods, research on algorithm learning has acquired a new dimension. Several architectures have appeared that are capable of learning algorithms of moderate complexity, such as sorting, addition or multiplication. The ultimate goal, of course, is to learn algorithms for tasks with unknown solutions. That is not viable yet; nevertheless, the techniques developed for algorithm learning may yield to progress in practical fields relevant now. For example, the Neural GPU architecture, which was designed primarily for algorithm learning, was recently extended to perform machine translation \cite{kaiser2016can}.

Neural GPU \cite{Kaiser2015NeuralGL} is the most promising among the proposed architectures for algorithm learning because it is the only one capable of learning multiplication that generalizing to inputs much longer than the training examples. However, it is fragile since only a tiny fraction of the trained models generalize well. 

In this paper, we study ways to improve the Neural GPU to obtain faster training and better generalization. The proposed improvements allow us to achieve substantial gains: the model can learn binary multiplication in 800 steps versus 30000 steps that are needed for the original Neural GPU, and, most importantly all the trained models generalize to 100 times longer inputs with less than 1\% error. The model can also learn a wider range of problems with similar generalization performance, e.g. the decimal multiplication, which is the first time it has been learned end-to-end. To learn decimal multiplication we use a different representation where each decimal digit is encoded in binary. 

The improvements that achieve these goals are removal of the parameter sharing relaxation, introduction of nonlinearities with saturation cost, introduction of a diagonal gating mechanism. We also improve the training schedule by training on all input lengths simultaneously and use a larger learning rate with AdaMax optimizer \cite{kingma2014adam}. We integrate gradient clipping into AdaMax. 

We analyze the impact of each improvement separately and show that all of them are relevant to the achieved performance. We find that using hard nonlinearities with saturation cost is the key factor to achieve good generalization.

\section{Related Work}

There are two primary approaches to algorithm learning $-$ recurrent networks and reinforcement learning. In the reinforcement learning approach, the algorithmic device called controller is decoupled from the environment where the program executes. The device operates in steps which consist of observing the current state of the environment and issuing commands which change the next state of it. Such approach (at least theoretically) scales well in time and memory domain, but elaborate techniques of training are necessary since the overall structure cannot be differentiated. Simple algorithms such as sequence copying and reversal can be learned with the current reinforcement learning techniques \cite{zaremba2015reinforcement, zaremba2016learning}. 

Recurrent networks have a simple computation cell that is unrolled in time according to the length of an input sequence. Such approach is employed by  LSTM\cite{hochreiter1997long} and GRU \cite{cho2014learning} networks for sequence classification. Thy scale with sequence length but each cell has a constant amount of memory that essentially limits the learnable problems to regular languages. 

There are several more elaborate architectures proposed for algorithm learning. \cite{graves2014neural} developed a Neural Turing Machine capable of learning and executing simple programs such as repeat copying, simple priority sorting, and associative recall. They use complicated memory addressing to make the model differentiable. 

\cite{joulin2015inferring} introduce differentiable stack and double linked list data structures. Pointer Networks \cite{vinyals2015pointer} use soft attention and generalize to a variable-sized output space depending on the input sequence length. This model was shown to be effective for combinatorial optimization problems such as the traveling salesman and Delaunay triangulation. Neural Random-Access Machines \cite{kurach2015neural} introduce a memory addressing scheme potentially allowing constant time access due to discretization. \cite{grefenstette2015learning} introduce more neural data structures and evaluate them on several sequence processing tasks. A hierarchical memory layout with logarithmic access time is introduced in \cite{andrychowicz2016learning} with both differentiable and reinforcement learning versions being presented. 

Grid LSTM \cite{kalchbrenner2015grid} allow explicit unrolling along time and memory dimension and are able to learn such tasks as addition and memorization. 


A different setting of algorithm learning is explored in \cite{reed2015neural} where a model is trained on execution traces instead of input and output pairs; this richer supervision allows to induce higher level programs. 

Neural GPU \cite{Kaiser2015NeuralGL} is the current state of the art in deep algorithm learning. It can learn fairly complicated algorithms such as addition and binary multiplication, but only a small fraction of the trained models generalize well. The authors train 729 models to find one that generalizes well. 
They have no success for training decimal multiplication. \cite{price2016extensions} is able to train Neural GPU on decimal multiplication by using curriculum learning when the same model is trained at first for binary multiplication then for base-4 and only then for decimal. 

The basic idea of Neural GPU architecture is promising, and in this paper, we will we work on making it more powerful.

\section{The Model}
Neural GPU was introduced by \cite{Kaiser2015NeuralGL}. It is a recurrent network with a multi-dimensional state where a Convolution Gated Recurrent Unit (CGRU) is applied to the state at every time-step. CGRU is a combination of convolution operation and GRU\cite{cho2014learning} which computes the state $s_t$ at time $t$ from the state at time $t-1$ according to the following rules:

\centerline{$\begin{aligned}
    u_t &= \sigma(U^{\prime}\ast s_{t-1}+B^{\prime}) \\
    r_t &= \sigma(U^{\prime\prime}\ast s_{t-1}+B^{\prime\prime})\\
    c_t &= \tanh(U\ast(r_t\odot s_{t-1})+B) \\
    s_t &= u_t\odot s_{t-1}+(1-u_t)\odot c_t
\end{aligned}$}

In the above equations, $U$, $U^{\prime}$, $U^{\prime\prime}$ are convolution kernel banks, $B$, $B^{\prime}$, $B^{\prime\prime}$ are bias vectors; these are the parameters that will be learned. $U \ast s$ denotes a convolution of a kernel bank $U$ with a state $s$; $u\odot s$ denotes element-wise vector multiplication and $\sigma$ is the sigmoid function.

Given an input of length $n$, it is embedded into the first state, each symbol independently, producing a state with its first dimension equal to $n$, then CGRU is applied to it several times, and output is read from the last state by using a softmax loss for each symbol. 

While keeping this general architecture, we introduce some changes that include both simplifications and enhancements. So we will give more details immediately regarding our new architecture while mentioning the points where it deviates from the original.  

\subsection{Simplifications}
The original architecture by Kaiser and Sutskever uses a 3-dimensional state with its third dimension being of fixed size equal to 4. This extra dimension is not essential for the network's performance. We use a simpler 2-dimensional state of shape [$n$,$m$] where $n$ is the length of input and $m$ is the number of maps. To match the 2-dimensional state, our convolution kernel banks are of shape [3,$m$,$m$]. We fix the filter length to 3. We confirmed experimentally that this is the optimal setting for all considered tasks.

We use $n$ applications of the convolutional unit, all with the same set of parameters. Original implementation uses $2n$ applications with two sets of parameters. Hence our network is less deep and contains fewer parameters to be learned. 

We do not use parameter sharing relaxation (6 parameter sets that were slowly pulled together during training). With enhancements described below, our model does not need this feature and the learning schedule becomes simpler. 

\subsection{Diagonal gates}
The gating mechanism incorporated in the CGRU facilitates data copying to the same cell in the next time-step. This is essential for bringing together features separated in time during training. However, for most tasks, it is also required to bring together features from both ends of the input. Therefore, we introduce gates that copy data to a neighboring cell in the next time-step. We call these diagonal gates. We split all maps of a state into 3 parts $s_{t}=(s_{t}^1, s_{t}^2, s_{t}^3)$. The first part has a gate from the same cell in the previous time-step as in a CGRU, the second part uses gate from the left neighbor cell, and the third part uses gate from the right neighbor cell. 

To implement the diagonal gates, we need to shift the parts $s_{t-1}^2$ and $s_{t-1}^3$ to the right and left respectively and then apply a CGRU to the result. Shifting can be conveniently expressed as a convolution. Right shift corresponds to convolution with filter [1,0,0], left shift to convolution with [0,0,1] and no shift to convolution with [0,1,0]. We define the Diagonal Convolutional Gated Recurrent Unit (DCGRU), which we will use instead of CGRU, as follows:

\centerline{
$\begin{aligned}
              s_t &= u_t\odot \tilde{s_t}+(1-u_t)\odot c_t \\
    \tilde{s}_{t} &= (\tilde{s}_{t}^1, \tilde{s}_{t}^2, \tilde{s}_{t}^3) \\
    \tilde{s}_t^1 &= s_{t-1}^1\ast [0,1,0] \\
    \tilde{s}_t^2 &= s_{t-1}^2\ast [1,0,0] \\
    \tilde{s}_t^3 &= s_{t-1}^3\ast [0,0,1]
\end{aligned}$} 

Definitions of $u_t$ and $c_t$ are the same as for CGRU. The division of maps into 3 parts is only conceptual; an implementation uses a depthwise convolution operating directly on $s_{t-1}$ convolving each map with the required convolution filter independently. 

\begin{figure}
  \begin{center}
    \centerline{
        \includegraphics[]{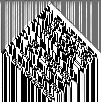}
        \includegraphics[]{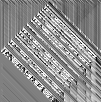}
        \includegraphics[]{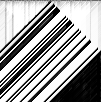}}
    \caption{Multiplication execution trace. A typical image from each of the parts $s^1$(left), $s^2$(middle), $s^3$(right) is shown.}
    \label{fig:mul-maps}
  \end{center}
\end{figure} 

To see that diagonal gates have impact, we can inspect execution trace of a trained model on some input. An execution trace is a collection of state values arising in the computation over all time steps. It is visualized as an image for each map where the input is given at the top, and the result is read from the bottom row of the image. 
In Fig~\ref{fig:mul-maps} we can see three maps of an execution trace performing binary multiplication on two 50 digit random numbers where each image is taken from each of the parts with a different gate direction. We can notice computing patterns that are aligned with the gate direction. The full set of maps is given in Appendix 1.  Fig~\ref{fig:sort-maps} shows maps from execution trace of a sorting task where 100 numbers in range 0 to 5 are sorted. These images look different than multiplication images, but patterns aligned with the gate direction are evident as well(see also Appendix 2 for the full set of maps). 

\begin{figure}
  \begin{center}
    \centerline{
        \includegraphics[]{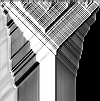}
        \includegraphics[]{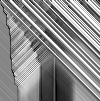}
        \includegraphics[]{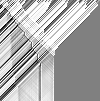}}
    \caption{Sorting execution trace. A typical image from each of the parts $s^1$(left), $s^2$(middle), $s^3$(right) is shown.}
    \label{fig:sort-maps}
  \end{center}
\end{figure}

Using gates that operate in different directions is not a novel idea. A similar mechanism is used in Grid LSTM \cite{kalchbrenner2015grid} where different units perform gating along different dimensions of the grid. But introduction such gates in a convolutional architecture is new.  

\subsection{Hard non-linearities}
\cite{Kaiser2015NeuralGL} have found that introducing gate cutoff improves performance. We go further and use hard\_tanh and hard\_sigmoid functions for all nonlinearities in the DCGRU. They are piecewise linear approximations of tanh and $\sigma$, namely

\centerline{$\begin{aligned}
    \mathrm{hard\_tanh}(x) &= \max(-1, \min(1, x)) \\
    \mathrm{hard\_}\sigma(x) &= \max(0, \min(1, (x+1)/2)) \text{\footnotemark} 
\end{aligned}$}
\footnotetext{Other literature may contain slightly different definition, but the following one is preferable in our case since we can use the same saturation cost for both functions.}

Hard nonlinearities train faster since their gradient does not approach zero. A drawback is the dying neuron problem that was observed for ReLU units where some neurons saturate for essentially all inputs. In the case of hard\_tanh and hard\_$\sigma$ units, this problem is even more pronounced since the range of the unit is bounded on both sides. Also, we use a relatively high learning rate that amplifies the problem. Therefore, we add an extra cost to the loss function to keep the units out of deadly saturation. For one unit we define 

\centerline{$\mathrm{saturation\_cost}(x)= \min(0,|x|-s\_limit)$}

with a parameter $s\_limit$ slightly less than 1 to keep the unit in its linear range. A value $s\_limit=0.9$ works well in our case. We calculate the saturation cost for each application of hard\_tanh($x$) and hard\_$\sigma(x)$, sum all of them together and add to the loss function with an appropriately small weight. We choose the weight such that the total saturation cost is 100 times smaller than the error loss.

Application of saturation cost increases the training time slightly, but training becomes more robust. The experimental analysis presented later in this paper shows that hard nonlinearities are essential to obtain good generalization and saturation cost allows to use larger learning rate safely. Note that we cannot use techniques like leaky ReLU \cite{maas2013rectifier},  parametric ReLU \cite{he2015delving} or exponential linear unit\cite{clevert2015fast} since they contain negative parts and the definition of GRU relies on values contained in their defined ranges. In particular, gate values have to be in the range [0,1] for gates to function properly. 

This technique of hard nonlinearities together with saturation cost can be applied to ordinary LSTM and GRU networks potentially yielding to improved performance. Also, it can be applied to ReLU networks, although it is not yet analyzed if it gives an advantage to the mentioned alternatives. 

\subsection{Training}
We train the model on inputs of all lengths simultaneously. As in the original architecture, we instantiate several bins of different lengths and place each training example into the smallest bin it fits and pad the remaining length. However, instead of training each bin separately, we sum their losses together and use one optimizer for the total loss. In this way, we avoid scheduling of bins and obtain faster convergence since, typically, several bins contribute to progress at each training step. 

We use a larger learning rate, i.e., $lr=0.005$ for a network with 96 maps; that is about 5 times larger than typically used. Training is not only faster but also more robust since the process can jump out of poor local minima. To keep learning converging, we use AdaMax optimizer \cite{kingma2014adam}. It has a strong guarantee that each parameter value will change by no more than $lr$ at each step. With more maps, we have to use a proportionally smaller learning rate since the sum over all maps contributes to each particular value in the next time step, and with more maps, the total contribution gets larger. We decrease the learning rate if no progress is made for 600 steps.

We integrate gradient clipping into AdaMax optimizer. We clip the gradient of each variable separately to the range proportional to its decayed maximum that is used internally by the optimizer. In this way, we do not need to set some predetermined clipping threshold. 
Gradient clipping is not strictly required for convergence since AdaMax optimizer is able to limit the step size even with high peaks in gradient. However, such high gradient sets a large decayed maximum and slows down further training. Clipping limits the growth of the decayed maximum and speeds up training. 

We use gradient noise of magnitude proportional to the learning rate. It was shown to give positive impact \cite{neelakantan2015adding}.

\subsection{Dropout}
Dropout improves training and generalization of the Neural GPU. To limit memory loss \cite{Kaiser2015NeuralGL} had to use dropout probability inversely proportional to the sequence length. 

We apply dropout only to the update vector $c_t$ of the CGRU as proposed in \cite{Semeniuta2016RecurrentDW}. That helps to avoid memory loss over many time steps. We chose a small constant dropout probability around 0.1. Practical experiments given in the next section confirms this choice as the best one.


\section{Evaluation}
We have implemented the proposed architecture in tensorflow. The code is available on GitHub\footnote{ \url{https://github.com/LUMII-Syslab/DNGPU}}. In this section, we compare the proposed architecture (denoted by DNGPU) with the original architecture (denoted NGPU) by \cite{Kaiser2015NeuralGL} as well as evaluate individual improvements of DNGPU proposed in this paper. 

We choose the binary multiplication task as the basis for the evaluation. It is the most complex task which can be learned by NGPU; more complex (of the studied tasks) being only decimal multiplication. Sorting, addition and other considered tasks can be learned more easily. 

We use NGPU implementation provided by the authors with settings proposed in the paper \cite{Kaiser2015NeuralGL}. We set the number of maps $nmaps=24$, dropout probability = 0.09. Other parameters we leave to default values used in the provided code.  

For DNGPU we use the number of maps $m=96$ to match the data amount carried in one state of NGPU (which use 24 maps in 4 rows). We use learning rate $lr= 0.05$, and dropout probability = 0.1. 

We use essentially the same training set as in the \cite{Kaiser2015NeuralGL} consisting of 10000 examples of every length up to 41 (two 20 bit numbers are multiplied). We trained 5 models with a random initialization and measured their accuracy on a test set containing random inputs of length 401 (two 200 bit numbers are multiplied). In this way, we can show both training speed and generalization in one graph. 
We used a computer with Intel Xeon E312 2.4 GHz processor, 64GB RAM and a Tesla K40 GPU card for testing.

\subsection{Performance and generalization}
\begin{figure}
    \includegraphics[width=\columnwidth]{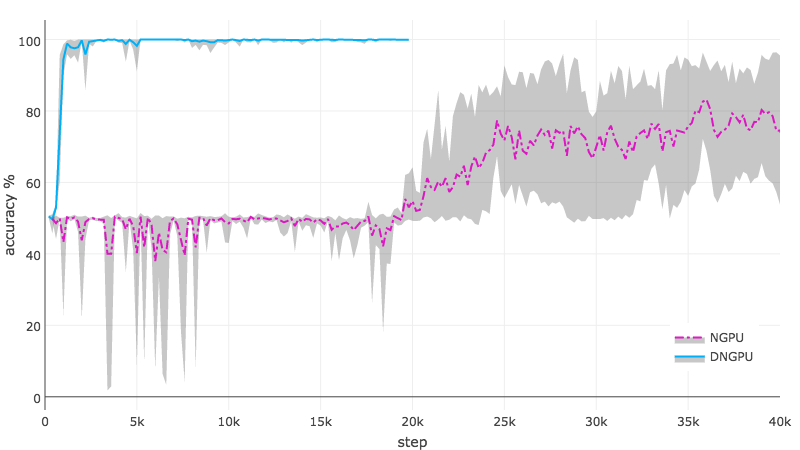}
    \caption{Accuracy on test set length 401 vs. step on binary multiplication.}
    \label{fig:NGPUvsDNGPU-BinMult41-iteration}
\end{figure}

To compare DNGPU with NGPU, we plot the accuracy of both models on the test for each step of training, see Fig~\ref{fig:NGPUvsDNGPU-BinMult41-iteration}. The solid lines show the average of all runs and the shaded area shows the scatter among different runs. Accuracy is defined as the percentage of correctly predicted output bits over all examples. We can see that the DNGPU converges much faster and achieves near 100\% accuracy in all runs. It requires only about 800 steps to reach 99\% accuracy. 

\begin{figure}
    \includegraphics[width=\columnwidth]{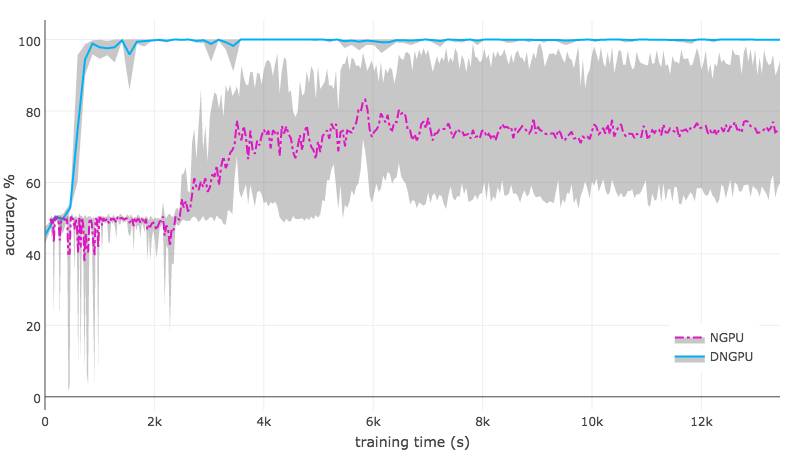}
    \caption{Accuracy on test set length 401 vs. training time on binary multiplication.}
    \label{fig:NGPUvsDNGPU-BinMult41-time}
\end{figure}

For a fair comparison, we have to take time per step into account. It is larger for our implementation since we evaluate gradient on all bins instead of one.  Fig~\ref{fig:NGPUvsDNGPU-BinMult41-time} shows the same data plot depending on the training time. DNGPU still has a significant advantage. Both implementations were run on the same hardware, and we excluded the time spent for accuracy evaluation. 

\begin{figure}[hb!]
    \includegraphics[width=\columnwidth]{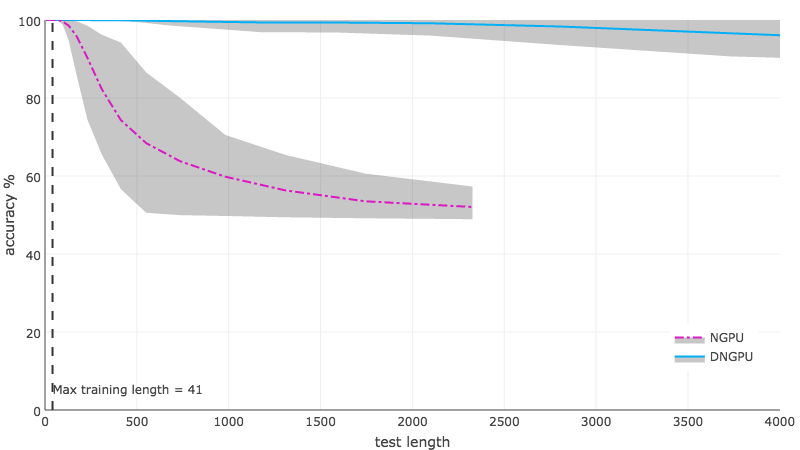}
    \caption{Accuracy on inputs of different lengths. The vertical dashed line shows the training length.}
    \label{fig:NGPUvsDNGPU-BinMult-generalization}
\end{figure}

To explore generalization beyond length 401, see Fig~\ref{fig:NGPUvsDNGPU-BinMult-generalization} which shows the accuracy of both architectures depending on input length. We see that DNGPU generalizes much better. All trained DNGPU models exceeded 90\% accuracy, and two out of five exceeded 99\% accuracy on length 4001. 

\begin{figure}
    \includegraphics[width=\columnwidth]{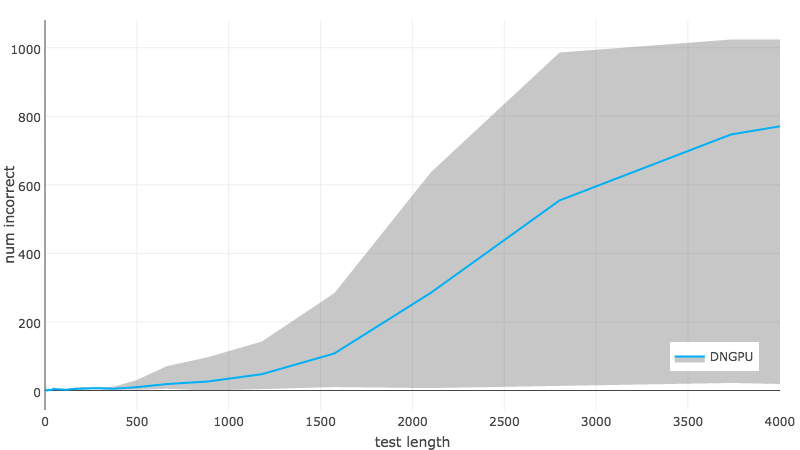}
    \caption{The number of test outputs containing at least one error.}
    \label{fig:DNGPU-BinMult-generalization-abs-wrong}
    \vspace{-1.5em}
\end{figure}

Although the accuracy achieved by DNGPU would be perfectly acceptable in other domain, demands for algorithm learning are higher. To speak that we have truly inferred an algorithm, it should generalize to arbitrary input length without a single error. This is not yet achieved. Fig~\ref{fig:DNGPU-BinMult-generalization-abs-wrong} shows the number of incorrectly predicted outputs (maximum value 1024) of the same trained models where we consider the output to be correct only if all its bits are correct. We can see that many of the longer examples contain errors (although the errors are few, the accuracy to be high). Moreover, most models have a few errors even in short examples, sometimes even on examples present in the training set. We do not know the exact causes for this, but some thoughts are given in \cite{price2016extensions}. Good news is that one model out of 5 gave only 19 outputs containing errors on length 4001, so by training more models it could be possible to find one which generalizes perfectly. 

To summarize, our architecture outperforms the original by a wide margin both in terms of training speed and of generalization. Our implementation consistently reaches 99\% accuracy on the test set of length 401 inputs in less than 15 minutes. The original NGPU trains slower and achieves 90\% accuracy only on some runs. Our findings about NGPU are consistent with a much more massive evaluation in \cite{neelakantan2015adding} Table 6 which shows that only a small fraction of its trained instances generalize well to length 401. Note that generalization of both models can be improved by increasing dropout probability together with the number of maps.

\subsection{Improvement impact analysis}

\begin{figure}[h]
    \includegraphics[width=\columnwidth]{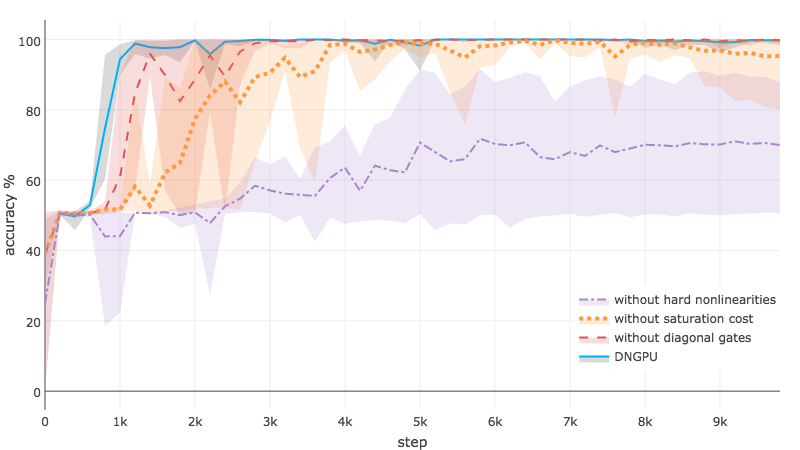}
    \caption{Impact of the proposed features. The magenta line shows the effect of using traditional soft tanh and sigmoid instead of hard ones. The yellow line shows performance with hard nonlinearities but without saturation cost. The red line shows performance without diagonal gates. The blue line is the suggested architecture.}
    \label{fig:ImpactAnalysisMisc}
    \vspace{-0.5em}
\end{figure}

We tried to understand how much each of the proposed enhancements contributes to the improved performance. Fig~\ref{fig:ImpactAnalysisMisc} shows how the model performs when one of the proposed features is turned off. A model trained without hard nonlinearities leads to especially poor performance. A closer look reveals that these models managed to fit the training set easily but generalized poorly to length 401. So, the key factor for achieving generalization is using hard nonlinearities. The same figure shows that hard nonlinearities without saturation cost also perform poorly. Training is unstable and does not converge to 100\% accuracy. Continuing training beyond 6000 steps slowly degrades the performance of the model. We can also see that without the diagonal gates the training becomes slower and more unstable. 

\begin{figure}
    \includegraphics[width=\columnwidth]{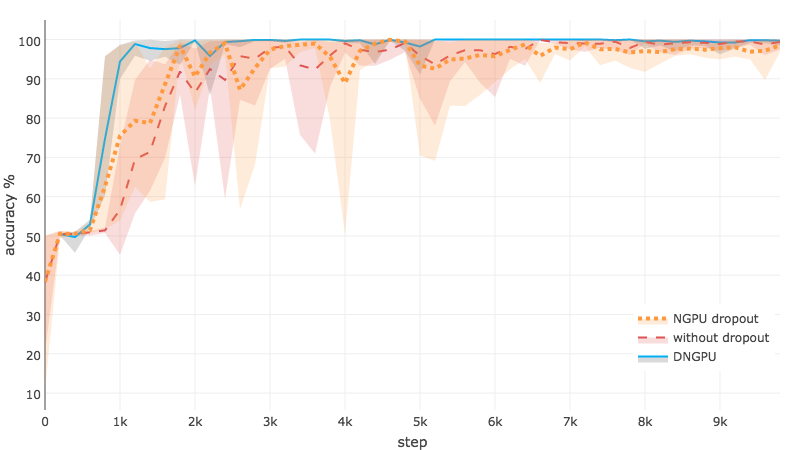}
    \caption{Performance of different dropout options. The blue line shows DNGPU performance which uses dropout by \cite{Semeniuta2016RecurrentDW}, the yellow line shows NGPU  dropout that was used in the code by Kaiser and Sutskever, the red line is without dropout.}
    \label{fig:ImpactAnalysisDropout}
    \vspace{-1.5em}
\end{figure}

In Fig~\ref{fig:ImpactAnalysisDropout} we analyze different dropout options. We can see that dropout suggested by \cite{Semeniuta2016RecurrentDW} performs better than the one employed in NGPU. Training without dropout is the worst option. 

\subsection{Decimal multiplication}
Our model can learn base-4 multiplication with consistently good generalization if we increase the number of maps to 192. However, like our predecessors, we did not succeed on the decimal multiplication task in its originally proposed form. 
But our architecture can learn decimal multiplication if we encode each decimal digit in binary. We use 4 bits per digit and mark the start of each digit with a different encoding of its first bit. Such encoding produces 4 times longer inputs and outputs. We implemented this encoding in input/output data generation part, but equivalently it can be implemented inside the Neural GPU itself by appropriate adjustment of its input and output layers. 

\begin{figure}
    \includegraphics[width=\columnwidth]{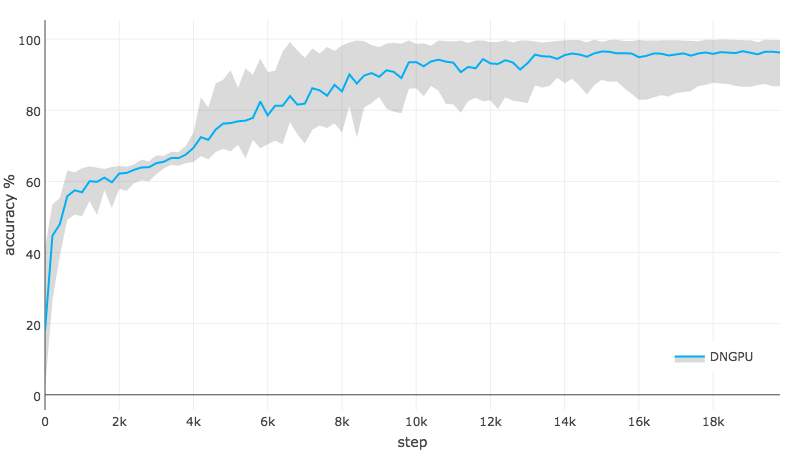}
    \caption{Accuracy on test set length 401 vs. step on decimal multiplication.}
    \label{fig:DNGPU-DecimalMult41-iteration}
\end{figure}

For evaluation, we increased the number of maps $m$ to 192 and performed training on examples of length 41 (multiplication of two 5 digit decimal numbers) and tested on examples of length 401 (multiplication of two 50 digit decimal numbers) as before. Fig~\ref{fig:DNGPU-DecimalMult41-iteration} shows the results. 

We were surprised to see that it generalized so well despite training only on very short examples containing two 5 digit numbers. Additionally, two models out of 5 generalized to length 401 with less than 1\% error. Of course, for better generalization we have to train on longer inputs. 

The binary encoding allows easier training. However, it comes with a significant overhead, i.e., a 4x increase in the input length which leads to a 16x increase in the unrolled model and a proportional increase training time and memory requirements.


\section{Conclusions}

We have presented several improvements to the Neural GPU architecture that substantially decrease training time and improve generalization. The main improvements are hard nonlinearities with saturation cost and a diagonal gating mechanism. We have shown that the hard nonlinearities with saturation cost contribute the most to obtaining better generalization. They may find further applications also in ordinary reccurent networks such as LSTM and GRU.

A larger learning rate together with AdaMax optimizer also helps the training performance, but the introduced saturation cost is essential to keep the learning convergent. 

The improved architecture can easily learn a variety of tasks including the binary multiplication on which other architectures struggle. If we increase the number of maps to 192, we can also learn base-4 multiplication with consistently good generalization. Furthermore, if we encode the decimal input/output digits in binary, the architecture can also learn decimal multiplication end-to-end. 

The improved architecture is considerably simpler than the original NeuralGPU, enabling an easier extension to handle harder problems. One such possible extension could be scaling the model to solve tasks requiring more than $n$ slots of memory or more than $n$ time steps. Simply enlarging the size of the model did not work well. So we leave the question of proper scaling of the model for future work. 

The correct generalization of the learned models to arbitrary large inputs is still an open problem, and it is not even clear why some models generalize, and others do not. With the proposed simpler model and faster training, it will be possible to address this question more effectively.


\section*{Acknowledgements}
We would like to thank the IMCS UL Scientific Cloud for the computing power and Leo Truk\v{s}\={a}ns for the technical support. The research was supported by ERDF project 1.1.1.1/16/A/135.


\bibliography{bibliography.bib}

\begin{thebibliography}{26}
\providecommand{\natexlab}[1]{#1}
\providecommand{\url}[1]{\texttt{#1}}
\expandafter\ifx\csname urlstyle\endcsname\relax
  \providecommand{\doi}[1]{doi: #1}\else
  \providecommand{\doi}{doi: \begingroup \urlstyle{rm}\Url}\fi

\bibitem[Amodei et~al.(2015)Amodei, Anubhai, Battenberg, Case, Casper,
  Catanzaro, Chen, Chrzanowski, Coates, Diamos, et~al.]{amodei2015deep}
Amodei, Dario, Anubhai, Rishita, Battenberg, Eric, Case, Carl, Casper, Jared,
  Catanzaro, Bryan, Chen, Jingdong, Chrzanowski, Mike, Coates, Adam, Diamos,
  Greg, et~al.
\newblock Deep speech 2: End-to-end speech recognition in english and mandarin.
\newblock \emph{arXiv preprint arXiv:1512.02595}, 2015.

\bibitem[Andrychowicz \& Kurach(2016)Andrychowicz and
  Kurach]{andrychowicz2016learning}
Andrychowicz, Marcin and Kurach, Karol.
\newblock Learning efficient algorithms with hierarchical attentive memory.
\newblock \emph{arXiv preprint arXiv:1602.03218}, 2016.

\bibitem[Angluin \& Smith(1983)Angluin and Smith]{angluin1983inductive}
Angluin, Dana and Smith, Carl~H.
\newblock Inductive inference: Theory and methods.
\newblock \emph{ACM Computing Surveys (CSUR)}, 15\penalty0 (3):\penalty0
  237--269, 1983.

\bibitem[Bahdanau et~al.(2014)Bahdanau, Cho, and Bengio]{bahdanau2014neural}
Bahdanau, Dzmitry, Cho, Kyunghyun, and Bengio, Yoshua.
\newblock Neural machine translation by jointly learning to align and
  translate.
\newblock \emph{arXiv preprint arXiv:1409.0473}, 2014.

\bibitem[Cho et~al.(2014)Cho, Van~Merri{\"e}nboer, Gulcehre, Bahdanau,
  Bougares, Schwenk, and Bengio]{cho2014learning}
Cho, Kyunghyun, Van~Merri{\"e}nboer, Bart, Gulcehre, Caglar, Bahdanau, Dzmitry,
  Bougares, Fethi, Schwenk, Holger, and Bengio, Yoshua.
\newblock Learning phrase representations using rnn encoder-decoder for
  statistical machine translation.
\newblock \emph{arXiv preprint arXiv:1406.1078}, 2014.

\bibitem[Clevert et~al.(2015)Clevert, Unterthiner, and
  Hochreiter]{clevert2015fast}
Clevert, Djork-Arn{\'e}, Unterthiner, Thomas, and Hochreiter, Sepp.
\newblock Fast and accurate deep network learning by exponential linear units
  (elus).
\newblock \emph{arXiv preprint arXiv:1511.07289}, 2015.

\bibitem[Gold(1967)]{gold1967language}
Gold, E~Mark.
\newblock Language identification in the limit.
\newblock \emph{Information and control}, 10\penalty0 (5):\penalty0 447--474,
  1967.

\bibitem[Graves et~al.(2014)Graves, Wayne, and Danihelka]{graves2014neural}
Graves, Alex, Wayne, Greg, and Danihelka, Ivo.
\newblock Neural turing machines.
\newblock \emph{arXiv preprint arXiv:1410.5401}, 2014.

\bibitem[Grefenstette et~al.(2015)Grefenstette, Hermann, Suleyman, and
  Blunsom]{grefenstette2015learning}
Grefenstette, Edward, Hermann, Karl~Moritz, Suleyman, Mustafa, and Blunsom,
  Phil.
\newblock Learning to transduce with unbounded memory.
\newblock In \emph{Advances in Neural Information Processing Systems}, pp.\
  1828--1836, 2015.

\bibitem[He et~al.(2015)He, Zhang, Ren, and Sun]{he2015delving}
He, Kaiming, Zhang, Xiangyu, Ren, Shaoqing, and Sun, Jian.
\newblock Delving deep into rectifiers: Surpassing human-level performance on
  imagenet classification.
\newblock In \emph{Proceedings of the IEEE international conference on computer
  vision}, pp.\  1026--1034, 2015.

\bibitem[Hochreiter \& Schmidhuber(1997)Hochreiter and
  Schmidhuber]{hochreiter1997long}
Hochreiter, Sepp and Schmidhuber, J{\"u}rgen.
\newblock Long short-term memory.
\newblock \emph{Neural computation}, 9\penalty0 (8):\penalty0 1735--1780, 1997.

\bibitem[Joulin \& Mikolov(2015)Joulin and Mikolov]{joulin2015inferring}
Joulin, Armand and Mikolov, Tomas.
\newblock Inferring algorithmic patterns with stack-augmented recurrent nets.
\newblock In \emph{Advances in neural information processing systems}, pp.\
  190--198, 2015.

\bibitem[Kaiser \& Bengio(2016)Kaiser and Bengio]{kaiser2016can}
Kaiser, {\L}ukasz and Bengio, Samy.
\newblock Can active memory replace attention?
\newblock In \emph{Advances in Neural Information Processing Systems}, pp.\
  3774--3782, 2016.

\bibitem[Kaiser \& Sutskever(2015)Kaiser and Sutskever]{Kaiser2015NeuralGL}
Kaiser, {\L}ukasz and Sutskever, Ilya.
\newblock Neural gpus learn algorithms.
\newblock \emph{arXiv preprint arXiv:1511.08228}, 2015.

\bibitem[Kalchbrenner et~al.(2015)Kalchbrenner, Danihelka, and
  Graves]{kalchbrenner2015grid}
Kalchbrenner, Nal, Danihelka, Ivo, and Graves, Alex.
\newblock Grid long short-term memory.
\newblock \emph{arXiv preprint arXiv:1507.01526}, 2015.

\bibitem[Kingma \& Ba(2014)Kingma and Ba]{kingma2014adam}
Kingma, Diederik and Ba, Jimmy.
\newblock Adam: A method for stochastic optimization.
\newblock \emph{arXiv preprint arXiv:1412.6980}, 2014.

\bibitem[Krizhevsky et~al.(2012)Krizhevsky, Sutskever, and
  Hinton]{krizhevsky2012imagenet}
Krizhevsky, Alex, Sutskever, Ilya, and Hinton, Geoffrey~E.
\newblock Imagenet classification with deep convolutional neural networks.
\newblock In \emph{Advances in neural information processing systems}, pp.\
  1097--1105, 2012.

\bibitem[Kurach et~al.(2016)Kurach, Andrychowicz, and
  Sutskever]{kurach2015neural}
Kurach, Karol, Andrychowicz, Marcin, and Sutskever, Ilya.
\newblock Neural random-access machines.
\newblock \emph{ERCIM News}, 2016, 2016.

\bibitem[Maas et~al.(2013)Maas, Hannun, and Ng]{maas2013rectifier}
Maas, Andrew~L, Hannun, Awni~Y, and Ng, Andrew~Y.
\newblock Rectifier nonlinearities improve neural network acoustic models.
\newblock In \emph{Proc. ICML}, volume~30, 2013.

\bibitem[Neelakantan et~al.(2015)Neelakantan, Vilnis, Le, Sutskever, Kaiser,
  Kurach, and Martens]{neelakantan2015adding}
Neelakantan, Arvind, Vilnis, Luke, Le, Quoc~V, Sutskever, Ilya, Kaiser, Lukasz,
  Kurach, Karol, and Martens, James.
\newblock Adding gradient noise improves learning for very deep networks.
\newblock \emph{arXiv preprint arXiv:1511.06807}, 2015.

\bibitem[Price et~al.(2016)Price, Zaremba, and Sutskever]{price2016extensions}
Price, Eric, Zaremba, Wojciech, and Sutskever, Ilya.
\newblock Extensions and limitations of the neural gpu.
\newblock \emph{arXiv preprint arXiv:1611.00736}, 2016.

\bibitem[Reed \& De~Freitas(2015)Reed and De~Freitas]{reed2015neural}
Reed, Scott and De~Freitas, Nando.
\newblock Neural programmer-interpreters.
\newblock \emph{arXiv preprint arXiv:1511.06279}, 2015.

\bibitem[Semeniuta et~al.(2016)Semeniuta, Severyn, and
  Barth]{Semeniuta2016RecurrentDW}
Semeniuta, Stanislau, Severyn, Aliaksei, and Barth, Erhardt.
\newblock Recurrent dropout without memory loss.
\newblock In \emph{COLING}, 2016.

\bibitem[Vinyals et~al.(2015)Vinyals, Fortunato, and
  Jaitly]{vinyals2015pointer}
Vinyals, Oriol, Fortunato, Meire, and Jaitly, Navdeep.
\newblock Pointer networks.
\newblock In \emph{Advances in Neural Information Processing Systems}, pp.\
  2692--2700, 2015.

\bibitem[Zaremba \& Sutskever(2015)Zaremba and
  Sutskever]{zaremba2015reinforcement}
Zaremba, Wojciech and Sutskever, Ilya.
\newblock Reinforcement learning neural turing machines-revised.
\newblock \emph{arXiv preprint arXiv:1505.00521}, 2015.

\bibitem[Zaremba et~al.(2016)Zaremba, Mikolov, Joulin, and
  Fergus]{zaremba2016learning}
Zaremba, Wojciech, Mikolov, Tomas, Joulin, Armand, and Fergus, Rob.
\newblock Learning simple algorithms from examples.
\newblock In \emph{Proceedings of the International Conference on Machine
  Learning}, 2016.

\end{thebibliography}
\bibliographystyle{icml2017}

\onecolumn
\section*{Appendix 1}
All 96 maps of an execution trace performing binary multiplication on two 50 digit random numbers. We can notice computing patterns that are aligned with the gate direction. Every 4 image rows correspond to maps with a different gate direction. 

\noindent\begin{minipage}{\textwidth}        
    \centering
    \includegraphics[height=0.9\vsize]{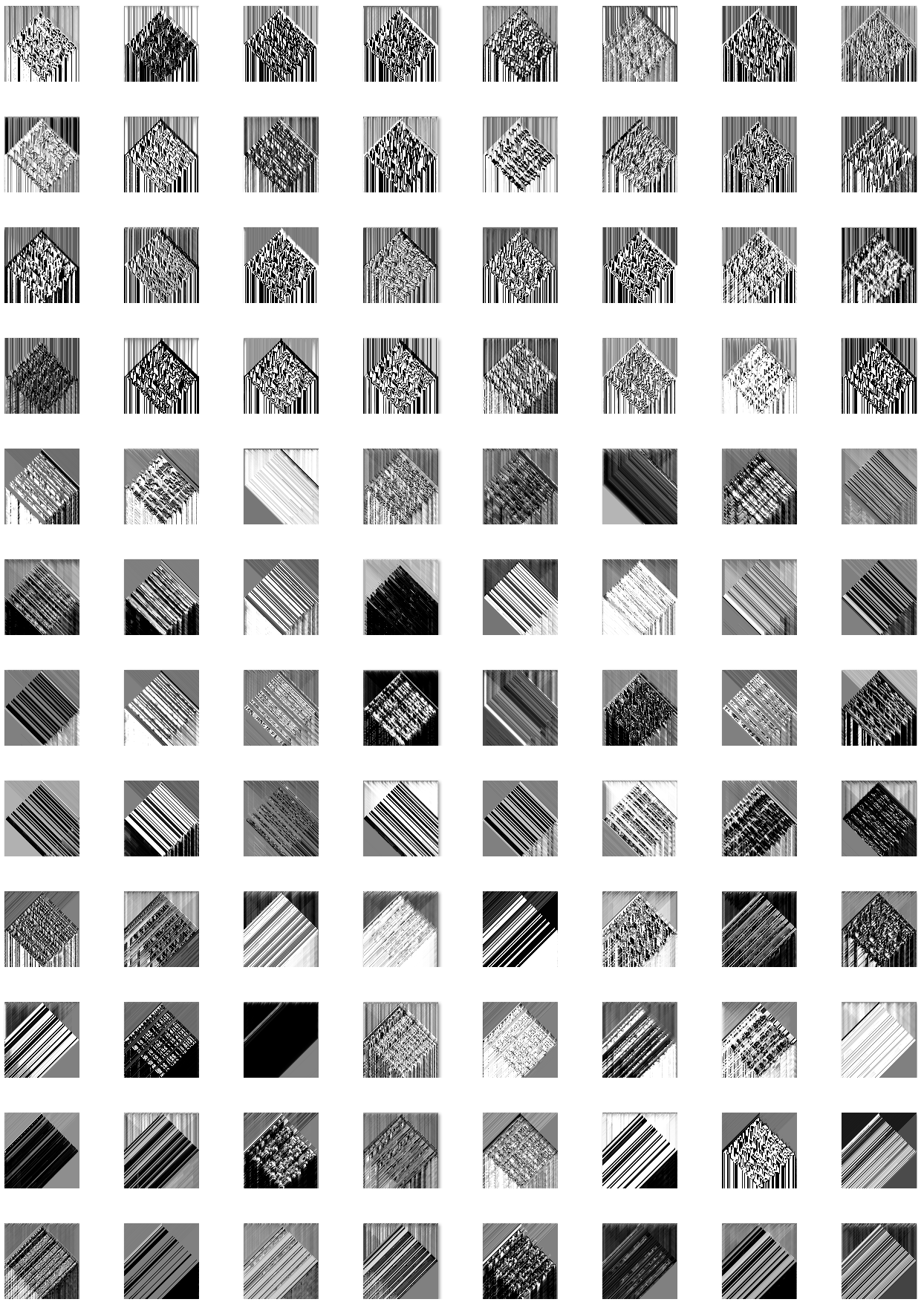}    
    
\end{minipage}

\section*{Appendix 2}
All 48 maps of an execution trace performing sorting where 100 numbers in range 0 to 5 are sorted. We can notice computing patterns that are aligned with the gate direction. Every 16 images correspond to maps with a different gate direction. 

\noindent\begin{minipage}{\textwidth}
    \centering
    \includegraphics[height=0.9\vsize]{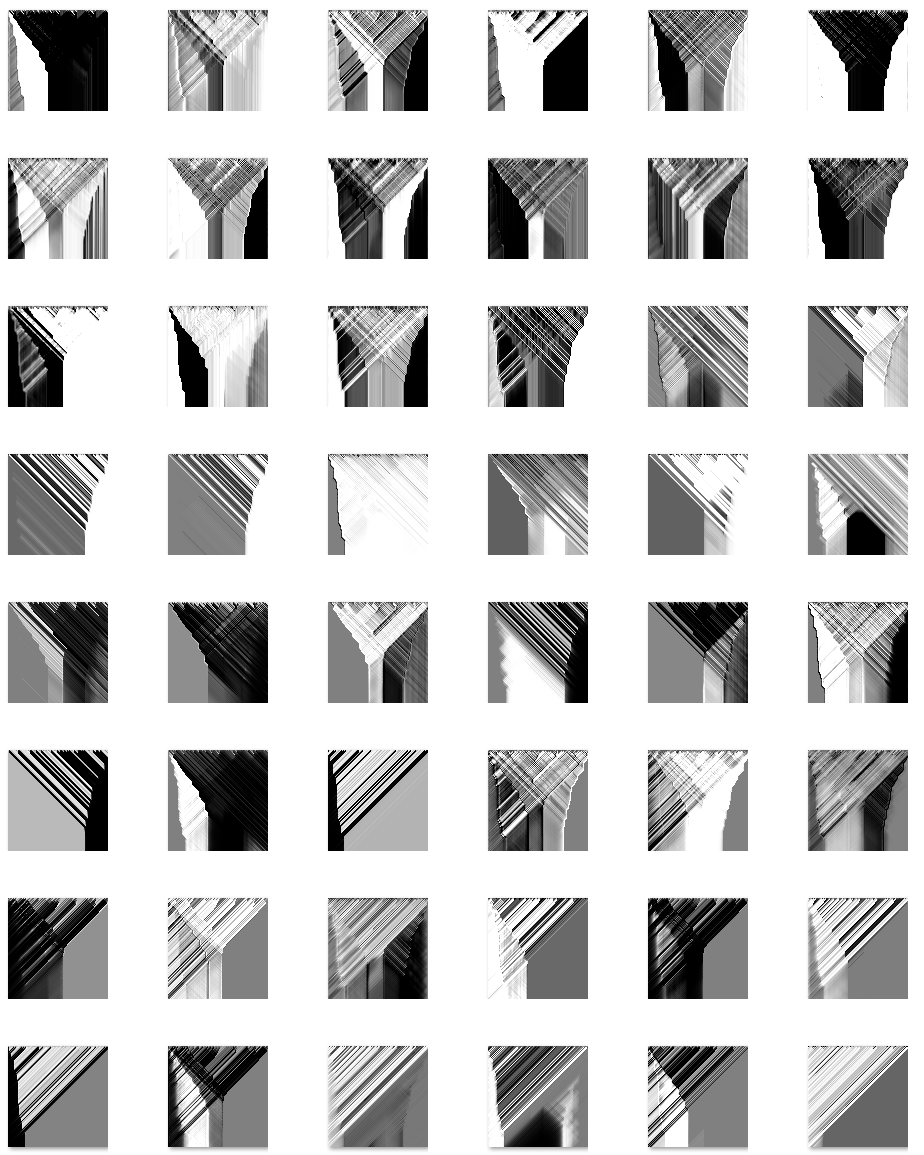}
    
\end{minipage}

\end{document}